\documentclass{article} 
\usepackage{iclr2027_conference,times}
\iclrfinalcopy

\usepackage{amsmath,amsfonts,bm}

\def\eqref#1{equation~\ref{#1}}

\def\1{\bm{1}}

\DeclareMathAlphabet{\mathsfit}{\encodingdefault}{\sfdefault}{m}{sl}
\SetMathAlphabet{\mathsfit}{bold}{\encodingdefault}{\sfdefault}{bx}{n}

\usepackage{amsthm}

\usepackage{amsmath}
\usepackage{amssymb}
\usepackage{algorithm}
\usepackage{algpseudocode}
\usepackage{booktabs}
\usepackage{float}
\usepackage{hyperref}
\usepackage{url}
\usepackage{booktabs}
\usepackage{tabularx}
\usepackage{wrapfig}
\usepackage{multirow}
\usepackage{rotating}
\usepackage{graphicx}
\usepackage{microtype}
\usepackage{xcolor}

\definecolor{colordel}{HTML}{a6a7a8}

\usepackage[capitalize,noabbrev]{cleveref}
\usepackage[normalem]{ulem}
\definecolor{myblue}{HTML}{598BE7}

\title{DRAM: Delta-rule Recurrent Associative Memory for Robot Manipulation Policies}

\author{
Xinyu Zhao~\thanks{Equal Contribution.}~~,~~Yixiang Shan$^{*}$, 
Tao Yang, Runyu Lei, Yiming Zhao, Jiaxin Fan, \\
\vspace{10pt}
\textbf{Zongbao Feng}\thanks{Corresponding Author.}~~~\& \textbf{Peng Jia} \\
Simplexity Robotics \\
\texttt{\{zhaoxinyu, shanyixiang, yangtao, leirunyu,  -, fanjiaxin,} \\
\texttt{fengzongbao, jiapeng\}@s-robots.com}
}

\begin{document}

\maketitle
\fancyhead[L]{\small Preprint.}

\begin{abstract}
    Robotic manipulation is inherently history-dependent, yet most pretrained robotic 
    policies condition on only the current observation or a short temporal window. Equipping such policies with long-term memory remains challenging: existing approaches either feed the backbone multi-frame observation windows, which substantially increase inference cost, or rely on pre-defined semantic features, which limit task generality and may also require the retraining of the backbone to adapt to the memory. We introduce DRAM (Delta-rule Recurrent Associative Memory), a plug-and-play memory module that can be attached to a wide range of pretrained robotic policies, endowing them with long-horizon memory without architectural modification or backbone retraining, requiring only task-specific post-training of the memory module and action expert. 
    DRAM maintains a fixed-size associative memory using gated delta-rule linear attention, with a modified update that incorporates all tokens within each frame in parallel. An architecture-agnostic readout integrates historical context into action prediction across different policy architectures.
    Experiments show that DRAM consistently improves frozen pretrained policies over short-context baselines and alternative compact memory designs, validating its effectiveness as a fixed-size, post-hoc memory module trained with the backbone frozen.
\end{abstract}
\vspace{-10pt}
\section{Introduction}
\vspace{-10pt}
Historical information is a fundamental requirement for robot
manipulation over extended horizons~\citep{shi2026memoryvla,li2026rememvla,jiang2026robottt}.
However, many manipulation tasks are only partially observable: the current observation
may not reveal objects that were previously visible but have since become
occluded, task stages that have already been completed, or the outcomes of
earlier interactions. Consequently, visually similar observations may
require different actions depending on the preceding history. 
Therefore, a policy must maintain and retrieve task-relevant history information across time to form
a more complete estimate of the environment state and task progress. 
However,
many existing robot policies condition their actions only on the current
observation or a short temporal window, preventing historical information outside that
window from influencing subsequent decisions and limiting their performance
on history-dependent tasks~\citep{li2026rememvla,jiang2026robottt}.

Recent studies have incorporated history as memory into robot policies. Some
retain compact recurrent summaries of past observations: RoboTTT compresses
visuomotor history into online-updated fast weights~\citep{jiang2026robottt},
and ReMem-VLA propagates temporal context through frame-level and chunk-level
recurrent queries~\citep{li2026rememvla}. Others store
historical features in an explicit memory bank and retrieve task-relevant
entries, as in MemoryVLA~\citep{shi2026memoryvla}. These
approaches differ substantially in how history is represented, updated, and
injected into the policy, and are instantiated as components of their
respective policy architectures. This motivates a memory mechanism that
maintains fine-grained historical information with bounded state size and
exposes the retrieved information through an architecture-agnostic interface.


Linear attention~\citep{choromanski2021performer} provides a natural foundation for this purpose; it replaces the softmax similarity with a factorizable kernel
and uses the associativity of matrix multiplication to aggregate
information before applying the queries~\citep{katharopoulos2020transformers}.
Under causal processing, this computation admits a recurrent implementation in
which historical associations are maintained in a fixed-size
matrix~\citep{katharopoulos2020transformers,schlag2021linear}.
Consequently, the size of compressed historical associations does not grow with the number of
preceding observations. However, the standard additive update continually
superposes new associations onto the existing state, making it difficult to remove previous associations
and can eventually cause interference between similar
keys~\citep{schlag2021linear,yang2024parallelizing}.

In this work, we introduce \textbf{DRAM}, a
\textbf{D}elta-rule \textbf{R}ecurrent \textbf{A}ssociative \textbf{M}emory
for robot manipulation policies. Building on delta-rule memory management~\citep{schlag2021linear,yang2024parallelizing}, instead of unconditionally adding each new association, DRAM first retrieves
the value currently associated with a key and then writes the residual between
this prediction and the new value. 
Therefore, the memory of DRAM is updated by correcting
its existing association rather than repeatedly accumulating the complete
value. This formulation enables DRAM to continuously incorporate new
observations into a fixed-size recurrent state while controlling redundant and
conflicting writes.

Moreover, DRAM operates on visual patch tokens while maintaining recurrent memory updates across frames. 
At each control step, all visual patches query the shared pre-update memory state carried over from preceding frames. Although they access the same memory snapshot, their distinct queries produce patch-specific historical features, which are fused with the current representations through a zero-initialized residual interface for action prediction.
Only after this read operation, the information of the current frame is written into memory using the delta rule, producing the state for the next frame. 
This \emph{read-before-write} design prevents the current observation from entering its own retrieval and avoids imposing an artificial sequential order among patches within the same image. 
Besides, implemented as a residual branch, DRAM preserves the original policy backbone and action-generation components, and can be integrated into different robot-policy backbones without architecture-specific redesign of their action decoders.
We evaluate DRAM on diverse robot manipulation tasks across multiple policy backbones, including real-robot experiments. The results show that DRAM consistently improves the performance of backbone policies.

Our contributions are summarized as follows:
\begin{itemize}
\item We introduce DRAM, a delta-rule recurrent associative memory for robot policies. DRAM maintains historical associations in a fixed-size state and updates them by correcting existing memory predictions rather than unconditionally accumulating new values.

\item We develop a patch-level, read-before-write memory mechanism that retrieves information from preceding frames before updating the recurrent state with the current observation, enabling fine-grained historical information to directly influence action generation.

\item We provide a lightweight residual interface that makes DRAM plug-and-play across different robot-policy backbones while preserving their original components, and demonstrate consistent improvements across diverse memory-dependent manipulation tasks, including real-robot evaluations.

\end{itemize}

\vspace{-10pt}
\section{Preliminaries}
\label{sec:preliminaries}
\vspace{-10pt}
\subsection{Problem Formulation}
\vspace{-10pt}
At environment step $t$, a robot receives a visual observation $o_t$,
a proprioceptive state $s_t$, and a task instruction $\ell$, and predicts
an action chunk $\mathbf{a}_t = a_{t:t+C-1}$ of length $C$.
Let $X_t$ denote the current-frame features produced by the policy backbone.
A policy without persistent memory predicts
\begin{equation}
    \mathbf{a}_t \sim \pi_\theta(\cdot \mid X_t, s_t, \ell).
\end{equation}
The current inputs may not reveal
task-relevant information from earlier interactions.
Memory augments this conditioning with a persistent representation
of the observation history.
\vspace{-7pt}
\subsection{Recurrent Memory}
\vspace{-7pt}
Recurrent memory carries a state $M_t$ across successive inputs:
\begin{equation}
    M_t = F_\psi(M_{t-1}, X_t),
    \label{eq:recurrent_memory}
\end{equation}
where $F_\psi$ integrates the current features with the previous state.
With a fixed-size state, the storage required for persistent memory
is independent of the number of preceding steps.

Linear attention provides an associative implementation of this
recurrence~\citep{katharopoulos2020transformers,schlag2021linear}.
For a generic token sequence, let
$q_i,k_i\in\mathbb{R}^{1\times d_k}$ and
$v_i\in\mathbb{R}^{1\times d_v}$, with feature maps absorbed into
the queries and keys.
Its unnormalized form admits
\begin{equation}
    S_i = S_{i-1} + k_i^\top v_i,
    \qquad
    r_i = q_i S_i,
    \label{eq:additive_memory}
\end{equation}
where $S_i\in\mathbb{R}^{d_k\times d_v}$ compresses historical
key--value associations.
This matrix-valued recurrence underlies the memory formulation
adopted in this work.
Other implementations include gated vector states in LSTMs and
memory tokens propagated through Transformer
layers~\citep{graves2012long,bulatov2022recurrent}.
Despite different state representations and update operators,
these mechanisms share the recursive form in
\eqref{eq:recurrent_memory}.
\vspace{-7pt}
\subsection{Delta-Rule Associative Memory}
\label{sec:delta-rule}
\vspace{-7pt}
The additive update in \eqref{eq:additive_memory} superposes
new associations without explicitly correcting existing ones.
DeltaNet instead updates memory using the error of its current
value prediction~\citep{schlag2021linear,yang2024parallelizing}.
For the local associative objective
$\mathcal{L}_i(S)=\frac{1}{2}\|k_iS-v_i\|_2^2$,
one gradient step writes the residual $v_i-k_iS$.
A gated formulation additionally applies data-dependent
retention~\citep{yang2025gated}:
\begin{equation}
    \begin{aligned}
        \bar S_{i-1} &= \gamma_i S_{i-1}, \\
        S_i &= \bar S_{i-1}
        + \beta_i k_i^\top
        \bigl(v_i-k_i\bar S_{i-1}\bigr),
    \end{aligned}
    \label{eq:gated_delta_memory}
\end{equation}
where $\gamma_i\in(0,1)$ controls retention and
$\beta_i\in[0,1]$ controls the correction strength,
with unit-normalized keys.
Unlike additive accumulation, this update modifies memory
according to the discrepancy between stored and incoming associations.
The standard recurrence processes tokens sequentially;
our method adapts this associative principle to joint updates
over the tokens of each environment frame.
\vspace{-7pt}
\section{Method}
\label{sec:method}
\vspace{-7pt}

\vspace{-10pt}
\begin{figure}[htbp]
    \centering
    \includegraphics[width=1.0\textwidth]{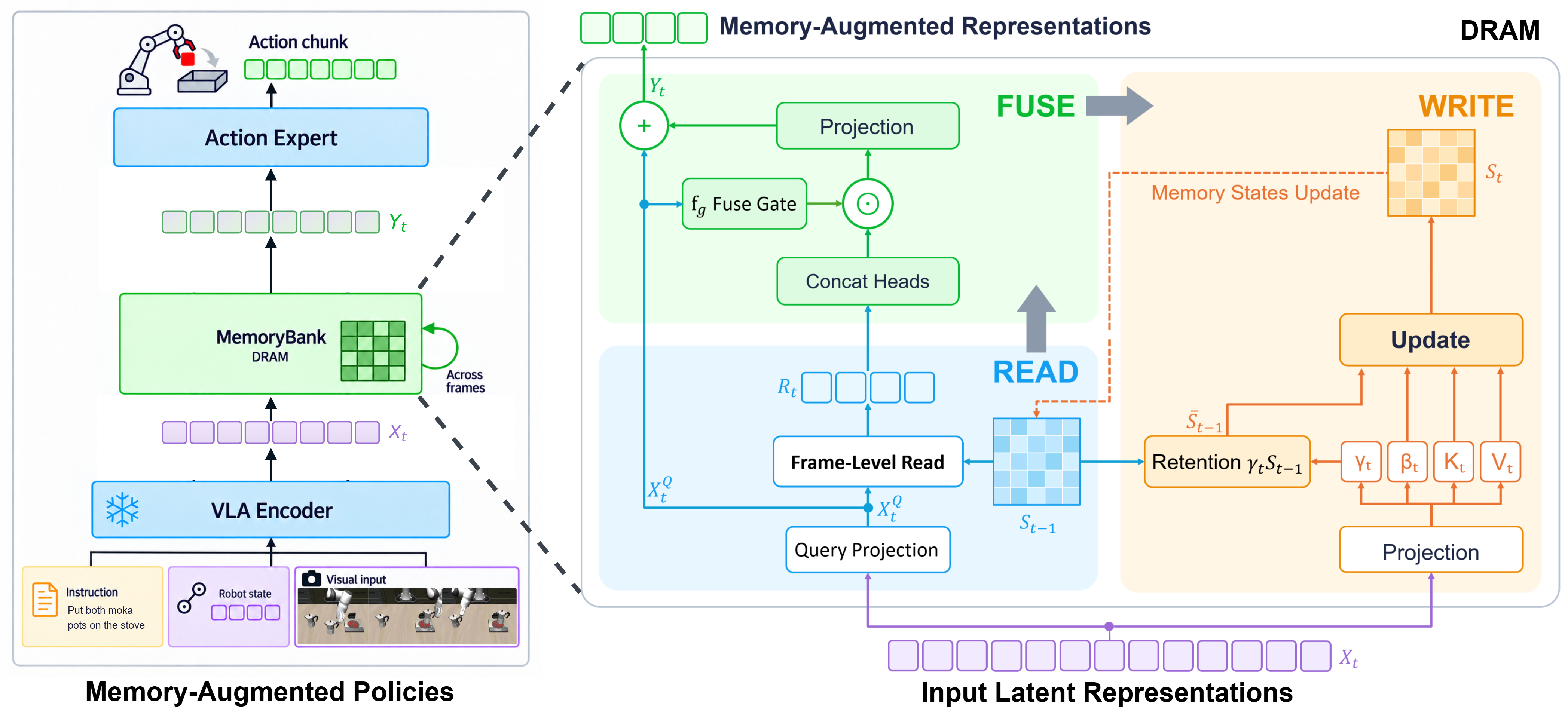}
    \vspace{-13pt}
    \caption{
Overview of DRAM.
\textbf{Left:} DRAM augments representations $X_t$ from a frozen VLA encoder with historical information, producing $Y_t$ for action prediction.
\textbf{Right:} The memory module follows a read--fuse--write procedure.
After reading, current-frame associations update the retained memory through a frame-level delta update, yielding $S_t$ for the next frame.
}
    \label{fig:DRAM}
\end{figure}
\vspace{-4pt}

We introduce the Delta-rule Recurrent Associative Memory (DRAM) module, a
frame-level recurrent memory for embodied policies that maintains a
fixed-size historical state while preserving the spatial structure of the
robot's visual representation and other observations. Our key principle is to
decouple the \emph{temporal granularity of memory recurrence} from the
\emph{spatial granularity of memory access}: each environment frame
constitutes one recurrent step, whereas individual visual patches
independently read from and write to a shared associative memory. At every
control step, the policy explicitly processes only the current observation,
while historical information is carried through a persistent Delta-rule-based
memory state. The module consists of three primitives:
\textsc{Read}, \textsc{Write}, and \textsc{Fuse}.

\vspace{-7pt}
\subsection{Framework Overview}
\vspace{-7pt}
\label{sec:memory_overview}
Figure~\ref{fig:DRAM} illustrates the framework of DRAM. At control step $t$, the
frozen VLM backbone encodes the current image, together with other inputs such
as the language instruction and the robot state, and exposes a collection of
frame-level latent token sequences, denoted collectively by $X_t = \{ X_t^Q, X_t^K, X_t^V \}$, whose
shapes and granularity are not fixed a priori. We retain the frame-level latent sequences rather than pooling
each observation into a single write vector---and we define the query and
write sources of one memory layer as views of these latents:
\begin{equation}
X^Q_t \in \mathbb{R}^{M\times D_Q},
\qquad
X^K_t \in \mathbb{R}^{N\times D_K},
\qquad
X^V_t \in \mathbb{R}^{N\times D_V},
\label{eq:sources}
\end{equation}
where $D_Q, D_K, D_V$ are the backbone feature dimensions. The three sources follow two interface patterns depending on the host
policy's architecture: a shared-source interface sets all three to the same
sequence ($M{=}N$), while a separate-source interface draws the write
sources $X^K_t, X^V_t$ from the backbone latents $X_t$ with a flexible
query source. We present the general case with $M$ query tokens and $N$
write tokens, and treat sharing as a special instance.

Given the normalized sources $Z^{\{\cdot\}}_t = \mathrm{LN}(X^{\{\cdot\}}_t)$,
the head-wise query, key, and value components are as follows:
\begin{equation}
Q_t^{(h)} = \mathrm{Norm}\big(Z^Q_t W_Q^{(h)}\big),
\qquad
K_t^{(h)} = \mathrm{Norm}\big(Z^K_t W_K^{(h)}\big),
\qquad
V_t^{(h)} = Z^V_t W_V^{(h)},
\label{eq:projections}
\end{equation}
where $h \in \{1,\dots,H\}$ indexes $H$ heads, $\mathrm{Norm}(\cdot)$ denotes
row-wise $\ell_2$ normalization, and
$Q_t \in \mathbb{R}^{H\times M\times d_k}$,
$K_t \in \mathbb{R}^{H\times N\times d_k}$,
$V_t \in \mathbb{R}^{H\times N\times d_v}$. The rows of $K_t$ and $V_t$ remain paired within each head. Crucially, their row index is \emph{not} an
environment timestamp: whether encoding spatial position, camera identity, or language order, all tokens of one frame belong to the same physical observation. They therefore form a single recurrent input rather than a sequence of timesteps---the premise of the frame-synchronous write in Section~\ref{sec:frame_joint_write}.

In the DRAM module, the historical information is carried by a set of $H$ associative memory states per layer,
\begin{equation}
\mathcal{S}_t = \big\{\, S_t^{(h)} \in \mathbb{R}^{d_k\times d_v}
\,\big|\, h = 1,\dots,H \,\big\},
\label{eq:state}
\end{equation}
whose size is independent of the episode length. Head indices and padding masks are omitted below except where heads interact. With these head-wise representations, the core of memory operation carries out three
primitives:
\begin{align}
\textsc{Read:}\quad  R_t &= \mathrm{Read}(Q_t,\, S_{t-1}),\label{eq:prim-read}\\
\textsc{Fuse:}\quad  Y_t &= \mathrm{Fuse}(X^Q_t,\, R_t),\label{eq:prim-fuse} \\[2.5pt]
\textsc{Write:}\quad S_t &= \mathrm{Write}(K_t,\, V_t,\, S_{t-1};\, \gamma_t\, \beta_t)\label{eq:prim-write}.
\end{align}
\textsc{Read} retrieves historical information for all queries in parallel
from the \emph{pre-update} state $S_{t-1}$. \textsc{Fuse} combines the readout $R_t$ with the query-source tokens to produce the
memory-augmented representation $Y_t$ consumed by the downstream policy.
\textsc{Write} then updates the state for the next frame: it applies the soft
retention gate $\gamma_t$ to the inherited state, then incorporates the
current key--value associations through a single joint update modulated by
$\beta_t$.

These equations specify dependencies rather than a mandatory serial schedule:
retrieval reads $S_{t-1}$ rather than the newly written $S_t$; retention
affects only the state-update branch; and fusion depends only on the query
source and the historical readout. The raw readout $R_t$ remains available
independently of fusion, allowing the policy adapter to decide how retrieved
information enters the host policy. We detail the readout path---\textsc{Read}
and \textsc{Fuse}---in Section~\ref{sec:read_path}, the state update---\textsc{Write}---in Section~\ref{sec:frame_joint_write}.

\vspace{-7pt}
\subsection{Memory Access and Historical Fusion}
\vspace{-7pt}
\label{sec:read_path}
At each control step, memory serves the policy before it is updated: the
current frame first retrieves history, and the retrieved content is fused into
the policy's token stream. 
\vspace{-10pt}
\paragraph{Read: frame-level retrieval from the pre-update state.}
Before any writing, every query token retrieves from the incoming state:
\begin{equation}
R_t^{(h)} = \frac{Q_t^{(h)} S_{t-1}^{(h)}}{\sqrt{d_k}} \in \mathbb{R}^{M\times d_v},
\qquad
R_t = \Big\Vert_{h=1}^{H} R_t^{(h)},
\label{eq:read}
\end{equation}
where $\Vert$ concatenates along the feature dimension. Queries are computed
from the current frame while $S_{t-1}$ contains only past frames, so retrieval
conditions on the current observation and task without the current frame
leaking into its own readout. 
\vspace{-7pt}
\paragraph{Fuse: zero-initialized gated residual.}
Fusion combines the concatenated readout $R_t$ with the query source through a
gated residual branch,
\begin{equation}
Y_t = \mathrm{Fuse}(X^Q_t, R_t)
= X^Q_t + f_O\!\Big(\sigma\!\big(f_G(X^Q_t)\big) \odot R_t\Big),
\label{eq:fuse}
\end{equation}
where $f_G:\mathbb{R}^{D_Q}\!\to\!\mathbb{R}^{Hd_v}$ is a row-wise
affine gate and $f_O:\mathbb{R}^{Hd_v}\!\to\!\mathbb{R}^{D_Q}$ is the
output projection. We zero-initialize $f_O$, making FUSE an exact
identity at initialization; the memory readout thus initially adds
no residual change to the query representations. The gate and
projection affect only the query stream, and an optional bidirectional
intra-frame mixer (attention and MLP) may further process $Y_t$
within that stream.
\vspace{-7pt}
\paragraph{From readout to action.}
How the fused representation is consumed depends on whether the query stream
shares its source with the write streams (Section~\ref{sec:memory_overview}).
With separate sources, the readout is a representation of its own: the query
slots' outputs enter the policy as additional context alongside an untouched host token stream (our $\pi_{0.5}$~\citep{intelligence2025pi05} host). With a shared source, history is instead carried by the token stream itself: \textsc{Fuse} returns the fused sequence $Y_t$ in place of the raw readout, memory layers stack as in an ordinary transformer with a depth independent of the host, and the action head reads $Y_t$ through its original interface with no modification (our DiT host). In both cases the downstream conditioning interface is unchanged. Appendix~\ref{app:integration} details the two patterns and their host
instantiations.

\vspace{-7pt}
\subsection{Frame-Synchronous Joint Write}
\label{sec:frame_joint_write}
\vspace{-7pt}
The original delta rule updates memory states one association at a time, a schedule dictated by the sequential nature of text. In embodied policies, however, all tokens of
one frame become available simultaneously, and their serialization, whether spatial layout, camera identity, or language order, carries no temporal meaning. 
We therefore promote the update unit from a single token to a whole frame. Each environment frame triggers exactly one memory transition, in which all of its key--value associations are written jointly.
\vspace{-7pt}
\paragraph{Gates: frame-level retention and token-level write strength.}
A scalar retention gate per head and frame softly decays the inherited state,
\begin{equation}
\gamma_t^{(h)} = \exp\!\big\{-\exp(a_h)\,
    \zeta(\bar g_t\, u_{\gamma,h} + b_{\gamma,h})\big\}\in(0,1),
\qquad
\bar S_{t-1}^{(h)} = \gamma_t^{(h)} S_{t-1}^{(h)},
\label{eq:retention}
\end{equation}
where $\bar g_t$ is the mask-aware mean of the write-source rows
(Section~\ref{sec:memory_overview}), $u_{\gamma,h}$ a learned direction, $a_h$ a learned scale, and
$\zeta(\cdot)$ the softplus function. A per-token write gate sets the strength of each association from the corresponding write-source row $g_{t, i}$:
\begin{equation}
\beta_{t,i}^{(h)} = \sigma(g_{t,i} u_{\beta,h} + b_{\beta,h}),
\quad
\hat\beta_{t,i}^{(h)} = \min(\beta_{t,i}^{(h)}, 0.99),
\quad
w_{t,i}^{(h)} = p_i\,\frac{\hat\beta_{t,i}^{(h)}}{1-\hat\beta_{t,i}^{(h)}},
\label{eq:write-gate}
\end{equation}
with $p_i\in\{0,1\}$ the padding mask. The odds reparameterization
$w=\hat\beta/(1-\hat\beta)$ is chosen so that, for a single unit-norm key, the
joint update below reduces exactly to the gated delta rule with strength
$\hat\beta$ (Appendix~\ref{app:single-token}); the cap bounds $w\le 99$ for
numerical conditioning (Appendix~\ref{app:numerics}).
\vspace{-7pt}
\paragraph{Update: frame-synchronous joint associative correction.}
The original delta rule is one gradient step on its prediction error
(Sec.~\ref{sec:delta-rule}). Instead of applying one such gradient step independently to each token in the latent sequence, we consider resolving all $N$ corrections simultaneously.

Suppressing head indices, let $\bar S=\gamma_t S_{t-1}$ and
$W_t=\mathrm{diag}(w_{t,1},\dots,w_{t,N})$. All $N$ associations of the current frame are written in one transition, defined as the minimizer of
\begin{equation}
S_t = \arg\min_{S\in\mathbb{R}^{d_k\times d_v}}\;
\frac{1}{2}\|S-\bar S\|_F^2
+ \frac{1}{2}\big\|W_t^{1/2}(K_t S - V_t)\big\|_F^2,
\label{eq:joint-objective}
\end{equation}
where $K_t\in\mathbb{R}^{N\times d_k}$ and $V_t\in\mathbb{R}^{N\times d_v}$ stack the keys and values of the $N$ write tokens row-wise, and
$W_t^{1/2}=\mathrm{diag}(\sqrt{w_{t,1}},\dots,\sqrt{w_{t, N}})$, so that the second term is simply the weighted sum $\frac12\sum_i w_{t, i}\|k_{t, i}S-v_{t, i}\|_2^2$. 
In the above equation, the first term anchors the update to retained history,
and the second asks one candidate state to explain \emph{all} current
key--value pairs simultaneously. Since the latter constrains only the
directions spanned by the current keys, the anchor---selecting the fitting
state closest to $\bar S$---keeps all unaddressed content intact.

Being quadratic in $S$, \cref{eq:joint-objective} is minimized where its
gradient vanishes, which yields the normal equation
$(I_{d_k} + K_t^{\top} W_t K_t)\, S_t = \bar S + K_t^{\top} W_t V_t$. 
Since $A \triangleq I_{d_k} + K_t^{\top} W_t K_t$ is symmetric positive
definite for any non-negative weights, the minimizer is unique with the closed-form solution:
\begin{equation}
S_t
= A^{-1}\big(\bar S + K_t^{\top} W_t V_t\big)
= A^{-1}\bar S + A^{-1}K_t^{\top} W_t V_t.
\label{eq:closed-form}
\end{equation}
All predictions $K_t\bar S$ are evaluated against the \emph{single} retained state, and $A^{-1}$ couples the residual corrections according to the key geometry. Detailed derivation can be found in Appendix~\ref{app:normal-eq}.

Note that the decay gate and the joint write are not independent forget
mechanisms: the historical term in \cref{eq:closed-form} is
$\gamma_t A^{-1} S_{t-1}$, so in directions constrained by current keys the
retained content is attenuated beyond the scalar decay---the joint fit
replaces it---while unconstrained directions undergo only soft decay.

In practice, $A^{-1}$ is never materialized: writing $S_t = \bar S + \Delta$,
the normal equation becomes
\begin{equation}
A\,\Delta = K_t^{\top} W_t\big(V_t - K_t\bar S\big)
\label{eq:correction}
\end{equation}which we solve by Cholesky factorization, returning $S_t = \bar S + \Delta$. Fully padded
frames leave $S_{t-1}$ untouched; valid frames with all-zero write weights return only the decayed $\bar S$. 
Masks, precision, and gradient flow through the solve are detailed in Appendix~\ref{app:numerics}.

\paragraph{Why not sequential token writes?} One could apply the delta rule token-by-token within a frame, iterating \cref{eq:gated_delta_memory} $N$ times. This chains $N$ recurrent states
indexed by an arbitrary serialization of the current frame: each correction
changes the state used to predict later values, so the final state depends on
the within-frame order even for identical inputs (a minimal counterexample is
in Appendix~\ref{app:normal-eq}). Chunked kernels accelerate this recurrence without removing
that dependence. We instead treat one frame as one simultaneous observation: \cref{eq:joint-objective} is the same local associative objective as the
delta rule, optimized jointly rather than greedily.

\vspace{-7pt}
\paragraph{Properties.}
Three properties matter for embodied use; all are proved in Appendix~\ref{app:derivation}. \emph{(i) Single-token correspondence.} For a single association $(k,v)$
with $\|k\|=1$ and weight $w=\hat\beta/(1-\hat\beta)$, \cref{eq:closed-form} reduces to the gated delta rule of \cref{eq:gated_delta_memory} with the clipped gate $\hat{\beta}$, so the joint-write rule strictly generalizes delta-rule memory management to simultaneous writes.
\emph{(ii) Within-frame permutation invariance.} Applying a common
permutation to the rows of $K_t,V_t$ and the diagonal entries of $W_t$
leaves $S_t$ unchanged in exact arithmetic: there is no serialization-induced recency bias among tokens of the same frame.
\emph{(iii) Bounded evidence accumulation.} $n$ repetitions of one association act with effective strength $nw/(1+nw)$: evidence accumulates
monotonically yet saturates below one, so repeated tokens reinforce rather
than dilute the write.


\vspace{-3pt}
\subsection{Training with Truncated Backpropagation through Time}
\label{sec:training}
\vspace{-7pt}
DRAM is designed for post-hoc integration on a pretrained policy: we train
the memory module, the policy-specific
adapter, and the action expert, while keeping the vision tower and the VLM
backbone fully frozen. This choice is deliberate rather than merely economical. The memory is trained jointly with the action expert through truncated BPTT (backpropagation through time). Co-training the deep VLM with the memory would chain gradients through the solve, the backbone layers, and the memory states of earlier frames---brittle to optimize and prone to destabilizing the pretrained representations the policy depends on.
The frozen backbone ensures the stability of the latent feature space, allowing memory to learn what to store, retrieve, and forget \emph{on top of frozen features}, with
all memory parameters optimized solely by the downstream control objective; we introduce no auxiliary reconstruction, retrieval, or memory-specific loss.
\vspace{-7pt}
\paragraph{Control objective.}
Any host policy objective can supervise the memory. In our setup, the action expert is trained using flow matching under its
data-to-noise convention: for a demonstrated action chunk $\mathbf{a}_t$, $\epsilon\sim\mathcal{N}(0,I)$, and flow time
$\tau = 0.999\,b + 0.001$ with $b\sim\mathrm{Beta}(1.5,1)$,
\begin{equation}
\mathbf{a}_t^{\tau} = (1-\tau)\,\mathbf{a}_t + \tau\,\epsilon,
\qquad
u_t = \epsilon - \mathbf{a}_t,
\qquad
\mathcal{L}_{\mathrm{FM}}
= \mathbb{E}\big\|\, v_{\theta,\psi}(\mathbf{a}_t^{\tau}, \tau;\, c_t, R_t) - u_t \,\big\|_2^2,
\label{eq:fm}
\end{equation}
using the host policy's padding-aware action reduction. Here $c_t$ denotes
the current-frame context, $\theta$ the action-expert parameters, and $\psi$
the memory and adapter parameters; $R_t$ collectively denotes the memory
readouts consumed by the adapter. Gradients of $\mathcal{L}_{\mathrm{FM}}$
flow through every memory operation---patch-level retrieval, gated writing,
and temporal retention---so the contents of memory are shaped directly by
their utility for control rather than by any hand-designed storage criterion.
\paragraph{Recurrent unrolling.}
Training unrolls consecutive frames within a TBPTT window. Inside a window,
gradients propagate through historical reads, the retention and write gates,
fusion where used, and the differentiable joint solve
(Appendix~\ref{app:derivation}); recurrent states are detached only at window
boundaries, not at activation-checkpoint segment boundaries. Valid-frame
losses are averaged over the contributing ranks and accumulated windows. At
deployment, each new observation triggers exactly one read--write transition,
the action expert denoises only when a new action chunk is required, and the
memory persists across chunks and resets across episodes. Note that a bounded
persistent state does not bound training-time activation memory, which still
grows with the TBPTT window length.

\vspace{-7pt}
\section{Experiments}
\label{sec:experiments}
\vspace{-7pt}


\subsection{Setup}
\label{sec:experimental_setup}
\vspace{-7pt}
\paragraph{Benchmark.}
Our main evaluation uses the four LIBERO suites~\citep{liu2023libero}:
LIBERO-Spatial, LIBERO-Object, LIBERO-Goal, and LIBERO-10.
These suites cover variations in spatial arrangements, objects,
and task goals, as well as longer manipulation tasks.
Note that LIBERO-10 provides a particularly relevant setting for examining performance over longer execution horizons.
\vspace{-7pt}
\paragraph{Compared methods.}
We compare DRAM with representative approaches that incorporate
historical information into robot policies:
MemoryVLA~\citep{shi2026memoryvla},
HAMLET~\citep{koo2026hamlet},
$\mu$VLA~\citep{cherepanov2026muvla},
and ContextVLA~\citep{jang2025contextvla}.
These methods employ explicit memory banks, historical token
aggregation, recurrent memory tokens, or compressed multi-frame
context.
For ContextVLA, we include its reported results with
$\pi_0$, $\pi_0$-FAST, and GR00T N1.5.
DRAM uses $\pi_{0.5}$ as its backbone policy in the main comparison.

\subsection{Main Results}
\label{sec:libero_main_results}
\vspace{-7pt}

\begin{table*}[t]
    \centering
    \caption{Success rates (\%) on LIBERO.
    Bold indicates the best result in each row, including ties.
    Superscripts indicate the backbone;
    DRAM uses $\pi_{0.5}$.
    Averages are recomputed over the four suites.
    Published results use their respective training and evaluation protocols.}
    \label{tab:libero_main}
    \resizebox{\textwidth}{!}{
    \begin{tabular}{lccccccc}
        \toprule
        Environment
        & MemoryVLA
        & HAMLET
        & $\mu$VLA
        & ContextVLA$^{\pi_0}$
        & ContextVLA$^{\pi_0\text{-FAST}}$
        & ContextVLA$^{\mathrm{G}}$
        & DRAM (Ours) \\
        \midrule
        LIBERO-Spatial
        & 98.40 & \textbf{99.00} & 93.00
        & 97.40 & 98.30 & 98.40 & \textbf{99.00} \\
        LIBERO-Object
        & 98.40 & \textbf{100.00} & 99.40
        & 98.20 & 99.20 & 99.00 & \textbf{100.00} \\
        LIBERO-Goal
        & 96.40 & \textbf{99.20} & 96.60
        & 96.40 & 95.60 & 97.20 & 98.00 \\
        LIBERO-10
        & 93.40 & 92.20 & 95.80
        & 93.80 & 90.20 & 93.40 & \textbf{99.00} \\
        \midrule
        Average
        & 96.65 & 97.60 & 96.20
        & 96.45 & 95.83 & 97.00 & \textbf{99.00} \\
        \bottomrule
    \end{tabular}
    }
    \vspace{-20pt}
\end{table*}

Table~\ref{tab:libero_main} summarizes the results.
DRAM achieves an average success rate of $99.00\%$,
the highest among the compared methods,
exceeding the next-best average of $97.60\%$ from HAMLET by $1.40$ percentage points.
The largest advantage appears on LIBERO-10, where DRAM reaches
$99.00\%$, exceeding the next-best result of $95.80\%$ from
$\mu$VLA by $3.20$ percentage points.
This result is consistent with the intended role of DRAM in
supporting decisions over longer task executions.
On LIBERO-Spatial and LIBERO-Object, DRAM achieves
$99.00\%$ and $100.00\%$, respectively, matching the best
reported results in the comparison.
On LIBERO-Goal, DRAM obtains $98.00\%$, below HAMLET's
$99.20\%$, indicating that its advantage is not uniform
across all suites.

These reported results indicate strong overall LIBERO performance.
Because the published methods use different training and evaluation
protocols, we next compare each backbone with and without DRAM
to assess its contribution under matched conditions.

\begin{wraptable}{r}{0.5\textwidth}
    \centering
    \caption{
        LIBERO success rates (\%) across policy backbones.
        Bold indicates the better result within each pair,
        including ties.
    }
    \label{tab:backbone_applicability}
    \resizebox{\linewidth}{!}{
    \begin{tabular}{lcccc}
        \toprule
        &
        \multicolumn{2}{c}{\textbf{DiT}} &
        \multicolumn{2}{c}{$\boldsymbol{\pi}_{0.5}$} \\
        \cmidrule(lr){2-3}
        \cmidrule(lr){4-5}
        \textbf{Suite}
        & Baseline & + DRAM
        & Baseline & + DRAM \\
        \midrule
        LIBERO-Spatial
        & 83.25 & \textbf{92.00}
        & 97.00 & \textbf{99.00} \\
        LIBERO-Object
        & 88.00 & \textbf{97.50}
        & 99.00 & \textbf{100.00} \\
        LIBERO-Goal
        & 91.00 & \textbf{98.00}
        & \textbf{98.00} & \textbf{98.00} \\
        LIBERO-10
        & 75.00 & \textbf{91.00}
        & 96.00 & \textbf{99.00} \\
        \midrule
        \textbf{Average}
        & 84.31 & \textbf{94.63}
        & 97.50 & \textbf{99.00} \\
        \bottomrule
    \end{tabular}
    }
    \vspace{-10pt}
\end{wraptable}

To examine whether DRAM benefits different policy architectures,
we compare each host policy with its DRAM-augmented counterpart
on LIBERO.
Note that this requires different integration patterns, which are detailed in Appendix~\ref{appendix:Flexible Integration Patterns}.
As shown in Table~\ref{tab:backbone_applicability}, DRAM improves
the DiT policy across all four suites, increasing its average
success rate from $84.31\%$ to $94.63\%$.
The largest improvement occurs on LIBERO-10, where success
increases from $75.00\%$ to $91.00\%$, a gain of $16.00$ percentage points.

DRAM also benefits the stronger $\pi_{0.5}$ backbone,
raising its average success rate from $97.50\%$ to $99.00\%$.
It improves performance on Spatial, Object, and LIBERO-10,
while maintaining the $98.00\%$ success rate on Goal.
Together, these results support the applicability of DRAM
to both evaluated backbones, with gains observed at different
levels of baseline performance.

\vspace{-4pt}
\begin{table}[H]
    \centering
    \caption{Success rates (\%) on RMBench.
    Bold indicates the best result in each row, including ties.
    $\pi_{0.5}$, X-VLA, and Mem-0 results are taken from the
    RMBench benchmark and use their respective training and
    evaluation protocols;
    DRAM uses the $\pi_{0.5}$ backbone with native-KV integration.}
    \label{tab:rmbench}
    \small
    \begin{tabular}{lcccc}
        \toprule
        Task & $\pi_{0.5}$ & X-VLA & Mem-0 & DRAM $\pi_{0.5}$ (Ours) \\
        \midrule
        Battery Try         & 16.00 & 26.00 & 28.00 & \textbf{65.00}  \\
        Blocks Ranking Try  & 6.00  & 1.00  & 18.00 & \textbf{45.00}  \\
        Cover Blocks        & 0.00  & 2.00  & 68.00 & \textbf{70.00}  \\
        Observe and Pick Up & 9.00  & 9.00  & 4.00  & \textbf{15.00}  \\
        Press Button        & 0.00  & 0.00  & 0.00  & \textbf{20.00}  \\
        Put Back Block      & 11.00 & 18.00 & \textbf{90.00} & 65.00 \\
        Rearrange Blocks    & 13.00 & 13.00 & 89.00 & \textbf{95.00}  \\
        Swap T              & 15.00 & 3.00  & 14.00 & \textbf{45.00}  \\
        Swap Blocks         & 24.00 & 16.00 & 67.00 & \textbf{100.00} \\
        \midrule
        Average             & 10.40 & 9.80 & 42.00 & \textbf{57.80} \\
        \bottomrule
    \end{tabular}
    \vspace{-12pt}
\end{table}

We further evaluate DRAM on RMBench~\cite{chen2026rmbench},
a bimanual manipulation benchmark in which task completion
explicitly requires episodic memory,
covering nine tasks that range from recalling a single past
event to accumulating evidence across multiple events.
Table~\ref{tab:rmbench} compares DRAM, instantiated on
$\pi_{0.5}$ with the native-KV integration pattern,
against the published $\pi_{0.5}$, X-VLA, and Mem-0 results
from the benchmark.
DRAM attains an average success rate of $57.8\%$, exceeding
the strongest baseline, Mem-0, by $15.8$ percentage points,
and achieves the highest success rate on eight of the nine tasks.
The gains are most pronounced on tasks that require retaining
episode-specific state, such as Battery Try ($65\%$ vs.\ $28\%$)
and Swap T ($45\%$ vs.\ $15\%$);
Put Back Block is the only task on which DRAM trails a baseline
($65\%$ vs.\ $90\%$ from Mem-0).
As the published results follow their respective training and
evaluation protocols, this comparison is indicative rather than
strictly controlled; nevertheless, it corroborates the LIBERO
findings on tasks where memory is indispensable. We evaluate DRAM on the $\pi_{0.5}$ backbone over $50$ rollouts
per task with the same training data budget as the benchmarked
policies ($50$ demonstrations per task); the baseline success
rates are taken directly from RMBench~\cite{chen2026rmbench},
which reports $100$ rollouts per task.

\subsection{Flexibility of Memory Integration}
\label{sec:exp:integration}
\vspace{-7pt}
DRAM separates recurrent memory operations from the policy
representations used to construct memory.
We examine this flexibility on $\pi_{0.5}$ through three
configurations that use different write sources while
retaining the same memory update rule and prefix-based
injection interface.
\vspace{-7pt}
\paragraph{Integration configurations.}
\emph{Native KV} uses the pre-RoPE keys and values from
each aligned VLM layer as memory write sources.
\emph{Hidden Tokens} derives memory keys and values
from the hidden tokens of the corresponding VLM layer.
\emph{Last Hidden} uses the final VLM hidden sequence
as the write source across memory layers and exposes
only the final memory readout.
In all configurations, the retrieved features are
projected into additional prefix keys and values
for the action expert.

\begin{table}[H]
    \centering
    \caption{
        Success rates (\%) of different DRAM configurations
        on LIBERO with $\pi_{0.5}$ from LeRobot.
        Averages are recomputed over the four displayed
        suite-level success rates.
        Bold indicates the best result in each row.
    }
    \label{tab:memory_integration}
    \resizebox{0.65\linewidth}{!}{
    \begin{tabular}{lcccc}
        \toprule
        & \textbf{Baseline}
        & \multicolumn{3}{c}{\textbf{DRAM (Ours)}} \\
        \cmidrule(lr){2-2}
        \cmidrule(lr){3-5}
        \textbf{Suite}
        & \textbf{$\pi_{0.5}$}
        & \textbf{Native KV}
        & \textbf{Hidden Tokens}
        & \textbf{Last Hidden} \\
        \midrule
        LIBERO-Spatial
        & 97.00 & \textbf{99.00} & 98.50 & 98.50 \\
        LIBERO-Object
        & 99.00 & \textbf{100.00} & 99.00 & 99.50 \\
        LIBERO-Goal
        & 98.00 & 98.00 & \textbf{99.50} & 96.00 \\
        LIBERO-10
        & 96.00 & \textbf{99.00} & 97.50 & 96.00 \\
        \midrule
        \textbf{Average}
        & 97.50 & \textbf{99.00} & 98.63 & 97.50 \\
        \bottomrule
    \end{tabular}
    }
    \vspace{-10pt}
\end{table}
\vspace{-10pt}
\paragraph{Results.}
As shown in Table~\ref{tab:memory_integration},
Native KV achieves the best average success rate of $99.00\%$,
exceeding the baseline by $1.50$ percentage points.
Hidden Tokens reaches $98.63\%$, while Last Hidden matches the
baseline average of $97.50\%$. Thus, the choice of write source
affects the benefit of DRAM even with the same memory update rule.

\subsection{Sensitivity to Memory Reset Interval}
\label{sec:memory_horizon}

To test sensitivity to interruptions of its recurrent state, we evaluate
DRAM on LIBERO-10 while resetting all memory states every $L$
observation-frame updates, with $L\in\{8,32,128,256\}$.
The full-memory setting resets only between episodes.
We use the same trained checkpoint and inference settings, with
$30$ rollouts per subtask ($300$ total).

\begin{figure}[H]
    \vspace{-4pt}
    \centering
    \includegraphics[width=0.5\linewidth]{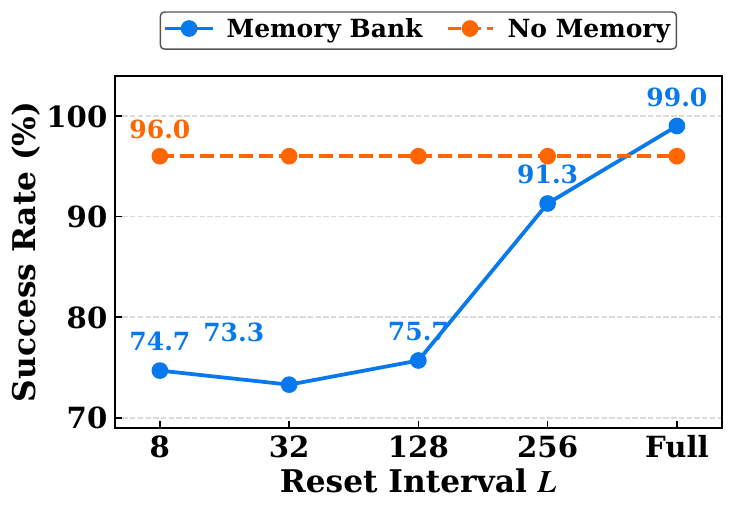}
    \caption{LIBERO-10 success versus memory reset interval $L$
    (in frames). Full resets only between episodes; the dashed line
    denotes the $\pi_{0.5}$ baseline without DRAM.}
    \label{fig:reset}
    \vspace{-10pt}
\end{figure}

As shown in Figure~\ref{fig:reset}, resetting memory every $8$, $32$,
or $128$ frames yields success rates of approximately
$73$--$76\%$, well below both full memory ($99.0\%$) and
the $\pi_{0.5}$ baseline without DRAM ($96.0\%$).
Increasing the interval to $256$ frames raises the success rate to
$91.3\%$, although it remains $7.7$ percentage points below full
memory and even still below the no-memory baseline.
These results show that the trained policy is sensitive
to interruptions of its recurrent state and that longer
intervals substantially reduce the performance loss. Periodic resets also alter the memory states encountered later in an episode, so these results characterize the policy's sensitivity to interrupted memory rather than establish a precise effective memory horizon.


\subsection{Real-world Experiments}
\label{sec:exp:real}
\begin{wraptable}{r}{0.4\textwidth}
    \vspace{-\intextsep}
    \centering
    \small
    \caption{Real-world success rates (\%).}
    \label{tab:realworld}
    \setlength{\tabcolsep}{5pt}
    \begin{tabular}{lcc}
        \toprule
        Task & DiT & DiT + DRAM \\
        \midrule
        Insertion & 90.0 & \textbf{100.0} \\
        Full Task & 0.0 & \textbf{85.0} \\
        \bottomrule
    \end{tabular}
\end{wraptable}

\textbf{Setup.}
Using the same DiT-based policy with cross-attention conditioning,
we evaluate insertion and a longer, multi-stage task with and without
DRAM. Each task-policy setting is evaluated in 20 trials; all other
training and evaluation settings are unchanged. Details are provided
in Appendix~\ref{appendix:setting of realworld}.

\textbf{Results.}
As shown in Table~\ref{tab:realworld}, DRAM raises insertion
success from 90.0\% to 100.0\% and full-task success from
0.0\% to 85.0\%. The larger gain on the multi-stage task
supports the value of persistent memory in this real-world setting.

%
\section{Related Work}
\label{sec:related_work}
\vspace{-7pt}
\subsection{Memory for Robotic Manipulation}
\vspace{-7pt}
Historical context helps robot policies resolve partial observability and track
task progress, and existing approaches differ in how they represent and
maintain it. MemoryVLA maintains an explicit bank of perceptual and cognitive
features, retrieved via cross-attention with redundant entries
consolidated~\citep{shi2026memoryvla}; HAMLET and ContextVLA instead compress
history into compact tokens---moment tokens aggregated over a history
window~\citep{koo2026hamlet}, or a single context token within the VLM
backbone~\citep{jang2025contextvla}. Others propagate compact recurrent states:
$\mu$VLA carries memory tokens through backbone
self-attention~\citep{cherepanov2026muvla}, ReMem-VLA maintains frame- and
chunk-level recurrent queries~\citep{li2026rememvla}, and MEM pairs short-term
visual context with a recurrent language summary of longer-term
semantics~\citep{torne2026mem}.

Closest to DRAM, RoboTTT scales visuomotor context through \emph{fast-weight
adaptation}~\citep{jiang2026robottt}, inserting test-time-training layers into
the action head that encode history by updating small networks during
execution. DRAM instead maintains an associative memory of token-level
key--value associations, retrieves historical features from the pre-update
state, and incorporates each frame through a joint memory update.
\vspace{-7pt}
\subsection{Recurrent Associative Memory}
\vspace{-7pt}
Linear attention admits a recurrent formulation that summarizes
historical key--value associations in a fixed-size
matrix~\citep{katharopoulos2020transformers}.
The connection to fast-weight learning provides an associative
interpretation of this state and motivates delta-rule updates
that correct existing predictions instead of repeatedly accumulating
complete values~\citep{schlag2021linear}.
Subsequent work develops efficient parallel algorithms for DeltaNet
and combines associative correction with data-dependent forgetting
in Gated DeltaNet~\citep{yang2024parallelizing,yang2025gated}.
These mechanisms provide the foundation for DRAM's memory operations.

Our focus is their use with structured visual observations, where tokens of
one frame are simultaneously available, and their serialization carries no
temporal meaning. DRAM therefore treats each frame as one recurrent input:
queries retrieve from the pre-update state, and current associations are
incorporated through a single joint optimization anchored to retained
history.


\vspace{-7pt}
\subsection{Frame-wise Delta Updates}
\label{sec:framewise-delta}
\vspace{-7pt}
Treating one frame as a simultaneous write unit has emerged independently in
video modeling. SANA-WM \citep{zhu2026sanawm} writes each frame into a
linear-attention world model explicitly, scaling keys by $1/\sqrt{U}$, while
a concurrent analysis \citep{zhu2026reflections} instead solves the per-frame
ridge problem, deriving essentially the same closed form and odds-based gate
conversion as DRAM. DRAM is concurrent and complementary: the joint write
lives in an external, policy-agnostic memory attached to a frozen policy,
remains strictly causal at control time, and is trained directly by the
control objective under TBPTT.



\vspace{-7pt}
\section{Conclusion}
\vspace{-7pt}
We presented DRAM, a recurrent associative memory module that augments
pretrained robot policies with persistent historical context. DRAM maintains
a fixed-size memory state with token-level associations, retrieves history
from the pre-update state, and incorporates each frame through a joint update
that imposes no sequential order among tokens of the same frame. Separating
memory operations from policy integration lets DRAM support different
representation interfaces while preserving the host policy's components, and
experiments on simulated and real-world tasks show consistent gains, most
pronounced on longer multi-stage tasks.


\section{Author Contributions}
\textbf{Xinyu Zhao} conceived the idea, led the experiments, and 
contributed to analyzing the experimental results, writing and revising the manuscript. \\
\textbf{Yixiang Shan} led the experimental design, the analysis of the experimental results, the writing of the initial 
draft, and contributed to revising the manuscript. \\
\textbf{Tao Yang, Runyu Lei, Yiming Zhao, and Jiaxin Fan} contributed to 
reviewing and revising the manuscript. \\
\textbf{Zongbao Feng and Peng Jia} supervised the research and contributed 
to reviewing and revising the manuscript.

\bibliography{iclr2027_conference}
\bibliographystyle{iclr2027_conference}

\appendix
\section{The Use of Large Language Models (LLMs)}
\label{app:host_policy}

We employ Large Language Models (LLMs) for grammar checking in our paper.

\section{Flexible Integration Patterns}
\label{app:integration}
\label{appendix:Flexible Integration Patterns}
DRAM admits the two integration patterns introduced in
Section~\ref{sec:read_path}: separate-source integration, where the query
stream is decoupled from the host's tokens, and shared-source integration,
where the host's frame tokens themselves carry the memory interface. The
choice is dictated by the host architecture rather than by the memory: the
\textsc{Read}, \textsc{Write}, and \textsc{Fuse} operations of
Section~\ref{sec:method} are identical in both. Within each pattern, three
further choices are orthogonal: where the write sources are tapped, how the
readout is delivered to the policy, and how memory layers align with host
layers. We instantiate the two patterns with the $\pi_{0.5}$ and DiT hosts
used in our experiments.

\subsection{Separate-source integration: the \texorpdfstring{$\pi_{0.5}$}{pi0.5} host}
\label{app:host-pi05}

With designated query slots (learned slots are the simplest case), the query
stream is decoupled from the host's token stream: the write sources are
designated backbone features of the current frame, while the query stream
alone evolves across stacked memory layers. The readout then enters the
policy as a separate representation---attention prefix slots, appended memory
tokens, or the pre-head representation---leaving the host's own token stream
untouched.

\begin{figure}[t]
    \centering
    \includegraphics[width=0.95\textwidth]{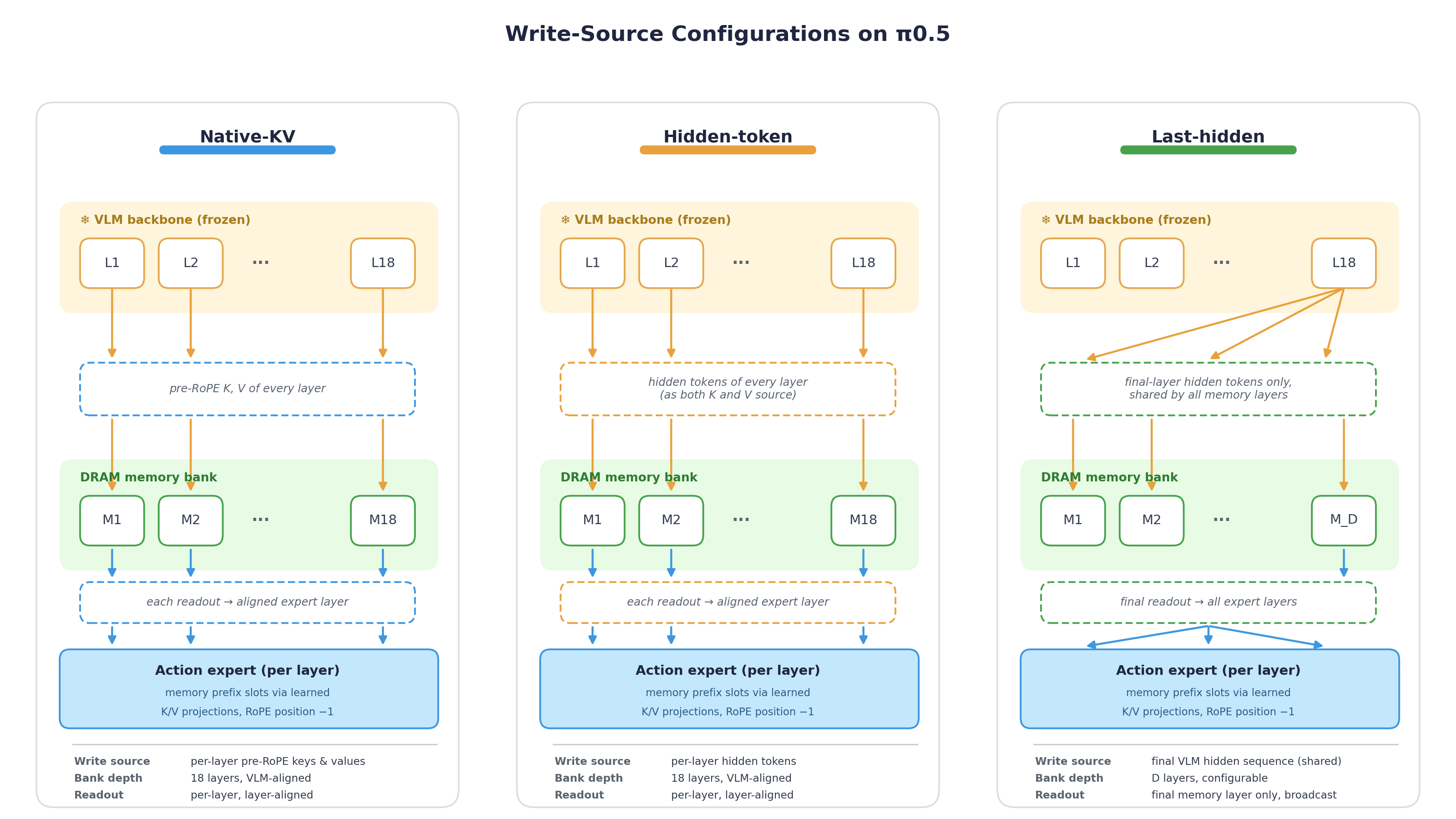}
    \caption{Write-source configurations for the $\pi_{0.5}$ host
    (separate-source integration). DRAM layers are aligned with the VLM
    backbone layers, and the write sources of memory layer $\ell$ are tapped
    from (i) the backbone attention's native key--value projections
    (\emph{Native KV}), (ii) intermediate hidden tokens (\emph{Hidden
    tokens}), or (iii) only the final-layer hidden states (\emph{Last
    hidden}). Query slots, readout injection, and the memory update are
    identical across configurations.}
    \label{fig:integration-modes}
\end{figure}

Our $\pi_{0.5}$ host follows this pattern. The frozen VLM backbone has 18
transformer layers, and DRAM layers are aligned with them: memory layer
$\ell$ reads and writes using features of backbone layer $\ell$, so
historical context is exchanged at matching representation depths. The query
stream is a set of $M$ learned query slots,
and the write sources $X^K_t, X^V_t$ are tapped from the backbone in one of
three configurations, illustrated in Figure~\ref{fig:integration-modes}:
(i) \emph{Native KV} reuses the backbone attention's own key--value
projections of each layer as the write sources, tapping representations the
backbone already computes;
(ii) \emph{Hidden tokens} tap the hidden token states of each backbone layer
and projects them with DRAM's own $W_K^{(h)}, W_V^{(h)}$;
(iii) \emph{Last hidden} uses the final-layer VLM hidden tokens as
a shared write source across memory layers, allowing the memory bank
depth to be chosen independently of the VLM depth; only the final
memory readout is passed to the action expert.
Section~4.3 compares the three configurations empirically (Table~3).

For the readout, the query slots retrieve from each memory layer,
and the fused readouts are injected as prefix slots into the action expert's
attention. The prefix slots use a reserved RoPE index ($-1$), for which
rotation of their keys is skipped; the action expert's original tokens
retain their position indices and thus their relative positions.
The zero-initialized output projection makes FUSE an identity at
initialization, so the memory readout initially adds no residual change
to the query representations.

\subsection{Shared-source integration: the DiT host}
\label{app:host-dit}

Without a deep multimodal backbone to tap, the current frame's own tokens
serve as the query, key, and value sources at once, and each memory layer
returns the fused sequence $Y_t^{(\ell)}$ in place of the raw readout,
feeding layer $\ell{+}1$ as in an ordinary transformer. Since history is
mixed into the tokens while preserving their structure, the action head
consumes $Y_t$ through its original interface with no modification; one
simply designates which layer's output feeds it.

Our DiT host follows this pattern, in the style of GR00T~N1.7: a vision tower
encodes all camera views of the current frame at once, and the resulting
patch tokens are flattened into a single sequence on which a DiT action head
conditions through cross-attention. The frame tokens serve as the query, key,
and value sources of DRAM ($M{=}N$); each memory layer applies the full
read--fuse--write procedure, optionally followed by the intra-frame mixer,
and memory layers stack with a depth chosen independently of the host.
Because history is carried by the tokens themselves---preserving their
number, ordering, and positional encoding---the DiT head reads the fused
sequence through its original cross-attention interface unchanged.

\paragraph{Practical principles.}
Two guidelines apply across both patterns. First, history-carrying tokens
should not disturb the host's positional scheme: injected readouts are marked
as positionless (a reserved RoPE index with unrotated values, as above) or
delivered through channels that bypass positional encoding altogether.
Second, readout granularity is orthogonal to the write-source choice:
layer-aligned readouts suit deep hosts whose representations evolve across
layers, whereas a single final-layer readout suffices for shallow hosts or
single-layer memory.





\section{Derivation and Properties of the Frame-Level Joint Write}
\label{app:derivation}

This appendix derives the closed form of the frame-synchronous joint write
(Section~\ref{sec:frame_joint_write}), proves the properties stated in the
main text, and details its numerical realization, including gradient flow
through the solve. Throughout, we suppress the frame index $t$ and the head
index $h$: $K\in\mathbb{R}^{N\times d_k}$ and $V\in\mathbb{R}^{N\times d_v}$
stack the unit-normalized keys and the values of the $N$ write tokens of one
frame row-wise, $k_i\in\mathbb{R}^{1\times d_k}$ and
$v_i\in\mathbb{R}^{1\times d_v}$ denote their $i$-th rows (so the prediction
for token $i$ is the row vector $k_iS$), $W=\mathrm{diag}(w_1,\dots,w_N)$
collects the non-negative write weights, and $\bar S=\gamma_t S_{t-1}$ is the
retained state.

\subsection{Normal equation, uniqueness, and closed forms}
\label{app:normal-eq}

\paragraph{Design of the joint objective.}
The single-token delta rule takes one \emph{explicit} gradient step on the local
prediction error $\mathcal{L}_i(S)=\frac12\|k_iS-v_i\|_2^2$, evaluated at the
retained state: $S=\bar S-\beta\,\nabla\mathcal{L}_i(\bar S)$. Applying it to
the $N$ tokens of a frame one after another chains $N$ such steps, each
evaluated at a different intermediate state, so the result depends on the
within-frame order: already with a scalar state, $\gamma{=}1$, $\beta{=}1$ and
the two tokens $(1,1)$ and $(1,0)$, sequential writes end at the \emph{last}
token's value ($0$ or $1$ depending on the order), whereas fitting both tokens
jointly yields $\tfrac12$ irrespective of their order. We therefore evaluate
all per-token errors against a \emph{single} candidate state and minimize them
jointly,
\begin{equation}
\mathcal{L}(S) \;=\; \frac12\sum_{i=1}^{N} w_i\,\|k_iS-v_i\|_2^2,
\label{eq:frame-loss}
\end{equation}
which asks one state to explain all current key--value pairs simultaneously.
By itself, however, $\mathcal{L}$ says nothing about the directions of the
state space that the current keys do not span: when $N<d_k$ the minimizer is
not unique, and the history stored in those directions could be altered
arbitrarily. We therefore add an anchor term and define the joint write as the
proximal step,
\begin{equation}
S \;=\; \arg\min_{S\in\mathbb{R}^{d_k\times d_v}}\;
\frac12\|S-\bar S\|_F^2 \;+\; \mathcal{L}(S),
\label{eq:joint-objective-app}
\end{equation}
This objective balances fitting the current key--value associations
against staying close to $\bar S$. Components not addressed by the
current keys remain unchanged, while addressed components are updated
according to the strength of the current evidence.
The anchor also fixes the relative weighting---with its weight set to $1$, each
$w_i$ plays the role of a per-token step size, and the odds choice
$w_i=\hat\beta_i/(1-\hat\beta_i)$ makes the single-token reduction reproduce the
gated delta rule with strength exactly $\hat\beta_i$
(Appendix~\ref{app:single-token}).

\paragraph{From sum to matrix form.}
We write the objective of \cref{eq:joint-objective-app}---identical to
\cref{eq:joint-objective} in Section~\ref{sec:frame_joint_write}---more
explicitly:
\begin{align}
J(S) &= \frac{1}{2}\|S - \bar S\|_F^2 + \frac{1}{2}\sum_{i=1}^{N} w_i \|k_i S - v_i\|_2^2 \notag\\
&= \frac{1}{2}\|S-\bar S\|_F^2
\;+\; \frac{1}{2}\left\|
\begin{bmatrix}
\sqrt{w_1}\,(k_1 S - v_1)\\
\vdots\\
\sqrt{w_N}\,(k_N S - v_N)
\end{bmatrix}
\right\|_F^2 \notag\\
&=\; \frac{1}{2}\|S-\bar S\|_F^2
\;+\; \frac{1}{2}\big\|W^{1/2}(KS - V)\big\|_F^2.
\label{eq:objective-expanded}
\end{align}
By the definition of the Frobenius norm:
the $i$-th row of $KS-V$ is the row vector $k_iS-v_i$, left-multiplying by
$W^{1/2}=\mathrm{diag}(\sqrt{w_1},\dots,\sqrt{w_N})$ scales row $i$ by
$\sqrt{w_i}$, and the squared Frobenius norm sums the per-row squared norms.
The matrix form of the second term is thus simply a compact notation for the
weighted sum of per-token prediction errors.

\paragraph{Gradient and normal equation.}
Since $W$ is symmetric, the objective expands as
\begin{align}
J(S) &=\; \frac{1}{2}\|S-\bar S\|_F^2
\;+\; \frac{1}{2}\big\|W^{1/2}(KS - V)\big\|_F^2 \notag \\
&=\; \frac12\operatorname{tr}\big((S-\bar S)^\top(S-\bar S)\big)
\;+\; \frac12\operatorname{tr}\big((KS-V)^\top W\,(KS-V)\big).
\label{eq:objective-trace}
\end{align}
To locate the minimizer, perturb $S$ along an arbitrary direction
$\Delta S\in\mathbb{R}^{d_k\times d_v}$ and expand in $\varepsilon$, using
$\frac12\|U+\varepsilon D\|_F^2=\frac12\|U\|_F^2+\varepsilon\langle
U,D\rangle_F+\frac{\varepsilon^2}{2}\|D\|_F^2$ on both terms:
\begin{align}
J(S+\varepsilon\Delta S)
&= J(S)
\;+\; \varepsilon\,\Big[\,
\big\langle S-\bar S,\,\Delta S\big\rangle_F
\,+\, \big\langle W^{1/2}(KS-V),\, W^{1/2}K\Delta S\big\rangle_F \,\Big] \notag\\
&\qquad\; +\; \frac{\varepsilon^2}{2}\,\Big[\,
\|\Delta S\|_F^2 \,+\, \big\|W^{1/2}K\Delta S\big\|_F^2 \,\Big] \notag\\
&= J(S)
\;+\; \varepsilon\,\big\langle (S-\bar S) + K^\top W(KS-V),\ \Delta S \big\rangle_F
\;+\; \frac{\varepsilon^2}{2}\,\big\langle \Delta S,\ A\,\Delta S \big\rangle_F,
\label{eq:perturbation}
\end{align}
where $\langle \cdot \rangle_F$ is the Frobenius inner product and $A\triangleq I_{d_k}+K^\top WK$. The second line of \cref{eq:perturbation} moves $W^{1/2}K$ across the inner product by the
adjoint rule $\langle X,\,W^{1/2}K\Delta S\rangle_F=\langle K^\top
W^{1/2}X,\,\Delta S\rangle_F$ (valid since $W$ is symmetric), and collects the
quadratic terms via $\|W^{1/2}K\Delta S\|_F^2=\langle\Delta S,\,K^\top
WK\Delta S\rangle_F$. Since the first-order coefficient holds for every
$\Delta S$, it identifies the gradient,
\begin{equation}
\nabla_S J(S) \;=\; (S - \bar S) \;+\; K^{\top} W (KS - V)
\;\in\; \mathbb{R}^{d_k\times d_v},
\label{eq:app-gradient}
\end{equation}
an affine map of $S$: the anchor contributes $S-\bar S$ and each token
contributes $w_i\,k_i^\top(k_iS-v_i)$, its prediction residual pulled back by
its key. Setting $\nabla_SJ(S)=0$ yields the normal equation
\begin{equation}
\underbrace{\big(I_{d_k} + K^{\top} W K\big)}_{A}\, S
\;=\; \bar S + K^{\top} W V .
\label{eq:app-normal}
\end{equation}
The second-order coefficient of \cref{eq:perturbation} is the Hessian
quadratic form, and it satisfies
\begin{equation}
\langle \Delta S,\,A\,\Delta S\rangle_F
\;=\; \|\Delta S\|_F^2 + \big\|W^{1/2}K\Delta S\big\|_F^2
\;\ge\; \|\Delta S\|_F^2 \;>\; 0
\qquad \text{for all } \Delta S\neq 0,
\label{eq:hessian-pd}
\end{equation}
because the Gram matrix $K^\top WK=\sum_{i=1}^N w_i\,k_i^\top k_i$ is positive
semidefinite for any non-negative weights. Hence $A\succeq I_{d_k}\succ 0$ for
\emph{any} keys---even repeated or rank-deficient ones---$J$ is strictly
convex, and the solution of \cref{eq:app-normal} is the unique global
minimizer. The identity supplied by the anchor is essential here: without it,
the fit term alone has no unique solution whenever the current keys fail to
span $\mathbb{R}^{d_k}$.

\paragraph{Closed forms.}
Since $A\succeq I_{d_k}$, it is invertible, and left-multiplying
\cref{eq:app-normal} by $A^{-1}$ gives the aggregate closed form. The
correction form follows from rewriting the retained state so that $A^{-1}$
cancels: by the definition $A \triangleq I_{d_k} + K^\top WK$,
\begin{equation}
A\bar S \;=\; (I_{d_k}+K^\top WK)\,\bar S \;=\; \bar S + K^\top WK\bar S,
\qquad\text{i.e.,}\qquad
\bar S \;=\; A\bar S - K^\top WK\bar S .
\label{eq:anchor-identity}
\end{equation}
Substituting this expression for $\bar S$ into the aggregate form lets
$A^{-1}$ act on $A\bar S$ and recover a bare $\bar S$:
\begin{align}
S
&= A^{-1}\big(\bar S + K^{\top} W V\big)
   &&\text{(aggregate form)} \notag\\
&= A^{-1}\big(A\bar S - K^{\top} W K \bar S + K^{\top} W V\big)
   &&\text{(substituting \cref{eq:anchor-identity})} \notag\\
&= \bar S + A^{-1} K^{\top} W \big(V - K\bar S\big)
   \;=:\; \bar S + \Delta .
   &&\text{(correction form)}
\label{eq:app-closed-form}
\end{align}
The aggregate form exposes how retained history and new evidence are blended by
$A^{-1}$; the correction form writes the update as the retained state plus a
coupled correction $\Delta$ on the prediction residual $V-K\bar S$, and is the
one we compute in practice (Appendix~\ref{app:numerics}). Equivalently,
$\Delta=-[\nabla^2 J]^{-1}\nabla J(\bar S)$ is the single Newton step from
$\bar S$ on the quadratic $J$, which lands exactly at the minimizer.

\subsection{Single-token correspondence}
\label{app:single-token}

For a single association $(k,v)$ with unit-normalized key $\|k\|_2=1$ and
weight $w$, the system matrix is a rank-one update of the identity,
$A = I_{d_k} + w\,k^{\top}k$, and the Sherman--Morrison formula gives
\begin{equation}
A^{-1} \;=\; I_{d_k} - \frac{w}{1+w\,k k^{\top}}\,k^{\top}k
\;=\; I_{d_k} - \frac{w}{1+w}\,k^{\top}k ,
\label{eq:app-sherman}
\end{equation}
since $kk^{\top}=\|k\|_2^2=1$. Applying this to the aggregate closed form,
\begin{align}
S
&= A^{-1}\big(\bar S + w\,k^{\top}v\big)
= \bar S - \frac{w}{1+w}\,k^{\top}(k\bar S)
+ w\,k^{\top}v - \frac{w^2}{1+w}\,k^{\top}(kk^{\top})v \nonumber\\
&= \bar S + \frac{w}{1+w}\,k^{\top}\big(v - k\bar S\big).
\label{eq:app-single-derivation}
\end{align}
The odds reparameterization $w = \hat\beta/(1-\hat\beta)$ of
\cref{eq:write-gate} is chosen precisely so that the effective
coefficient collapses to the gate itself:
\begin{equation}
\frac{w}{1+w} \;=\; \hat\beta
\qquad\Longrightarrow\qquad
S \;=\; \bar S + \hat\beta\, k^{\top}\big(v - k\bar S\big)
\;=\; \gamma_t S_{t-1}
+ \hat\beta\, k^{\top}\big(v - \gamma_t k S_{t-1}\big),
\label{eq:app-single-token}
\end{equation}
which is exactly the gated delta rule of \cref{eq:gated_delta_memory} with
correction strength $\hat\beta$. The frame-synchronous joint write therefore
strictly generalizes the gated delta rule: the latter is the $N=1$ special
case of the former. The same equivalence between an odds-parameterized gate
and the rank-one closed form is derived concurrently in the analysis of
\citet{zhu2026reflections}. The cap $\hat\beta\le 0.99$ bounds
$w\le 99$, keeping $A$ well conditioned (Appendix~\ref{app:numerics}).

\subsection{Permutation invariance and bounded accumulation}
\label{app:properties}

\paragraph{Permutation invariance (Property ii).}
Let $\Pi\in\{0,1\}^{N\times N}$ be a permutation matrix and consider the
permuted inputs $K'=\Pi K$, $V'=\Pi V$, and $W'=\Pi W\Pi^{\top}$ (the
diagonal entries of $W$, including the padding mask carried by
$w_i=p_i\hat\beta_i/(1-\hat\beta_i)$, are permuted along with the rows).
Both sides of the normal equation are unchanged:
\begin{equation}
K'^{\top} W' K' = K^{\top}\Pi^{\top}\Pi W\Pi^{\top}\Pi K = K^{\top} W K,
\qquad
K'^{\top} W' V' = K^{\top} W V,
\label{eq:app-permutation}
\end{equation}
since $\Pi^{\top}\Pi=I_N$. Hence $A'=A$; the right-hand side is identical,
and the unique minimizer $S$ is the same in exact arithmetic. No ordering of
the tokens within a frame---spatial raster, camera identity, or language
order---can influence the written state.

\paragraph{Bounded evidence accumulation (Property iii).}
Suppose one association $(k,v)$ with $\|k\|_2=1$ and weight $w$ appears $n$
times in a frame. The Gram matrix is $K^{\top}WK = nw\,k^{\top}k$, and
Sherman--Morrison as in \cref{eq:app-sherman}--\cref{eq:app-single-derivation} gives
\begin{equation}
S \;=\; \bar S + \frac{nw}{1+nw}\,k^{\top}\big(v - k\bar S\big).
\label{eq:app-bounded}
\end{equation}
The effective correction strength $nw/(1+nw)$ increases strictly with
$n$ for $w>0$, approaching $1$ from below. Repeated evidence therefore
\emph{reinforces} the write---unlike token-count averaging, which would
keep the strength at $w/(1+w)$ regardless of $n$---while the correction
strength remains below $1$ for every finite $n$.

\subsection{Numerical realization: masking, precision, and gradient flow}
\label{app:numerics}

\paragraph{Masking.}
Padded write tokens carry $p_i=0$, hence $w_i=0$: they contribute neither to
the Gram matrix nor to the right-hand side, and the mask-aware statistics of
the gates ($\bar g_t$) use the same mask. Fully padded frames skip the update
entirely, leaving $S_{t-1}$ untouched (in particular, no decay is applied);
valid frames whose write weights are all zero give $A=I_{d_k}$ and return
only the decayed state $S_t=\bar S$.

\paragraph{Conditioning and the role of the cap.}
With unit-normalized keys,
$\lambda_{\max}(K^{\top}WK) \le \sum_{i=1}^{N} w_i\,\|k_i\|_2^2
= \sum_{i=1}^{N} w_i \le 99\,N$,
while the anchor guarantees $\lambda_{\min}(A)\ge 1$. Hence
\begin{equation}
\kappa(A) \;\le\; 1 + 99\,N ,
\label{eq:app-conditioning}
\end{equation}
linear in the number of write tokens and independent of the data. The cap
$\hat\beta\le 0.99$ is what makes this bound possible, and the identity from
the anchor term is what makes the lower bound data-independent---the
numerical counterpart of its uniqueness role in
Appendix~\ref{app:normal-eq}.

\paragraph{Forward solve.}
We never materialize $A^{-1}$. Writing $S=\bar S+\Delta$, the normal equation
becomes $A\,\Delta = K^{\top}W\big(V - K\bar S\big)$, which we solve by
Cholesky factorization $A=LL^{\top}$ per frame, head, and memory layer; the
same factorization is reused for all $d_v$ columns of the right-hand side.
The Gram matrix and the solve are computed in FP32 with autocast and TF32
disabled, and the persistent states are stored in FP32; the rest of the
network is unaffected. The per-frame cost is $(O(Nd_k^2+Nd_kd_v+d_k^3+d_k^2d_v)$ per head, independent of episode length.

\paragraph{Gradient flow.}
The solve is differentiable, and we backpropagate through it by implicit
differentiation rather than through the Cholesky operator. Given the upstream
gradient $G=\partial\mathcal{L}/\partial S\in\mathbb{R}^{d_k\times d_v}$ at
the converged solution, the adjoint variable $\Lambda$ solves the transposed
system---which is the same system, since $A$ is symmetric---reusing the
stored factorization:
\begin{equation}
A\,\Lambda = G .
\label{eq:app-adjoint}
\end{equation}
The input gradients then follow from differentiating
$AS = \bar S + K^{\top}WV$:
\begin{align}
\frac{\partial\mathcal{L}}{\partial \bar S} &= \Lambda,
\qquad
\frac{\partial\mathcal{L}}{\partial V} = W K \Lambda,
\qquad
\frac{\partial\mathcal{L}}{\partial w_i}
= \big(K\Lambda\big)_i \big(V - KS\big)_i^{\top},
\label{eq:app-grads}\\
\frac{\partial\mathcal{L}}{\partial K}
&= W\big(V - KS\big)\Lambda^{\top} \;-\; W K\,\Lambda S^{\top},
\nonumber
\end{align}
where $(\cdot)_i$ denotes the $i$-th row. The weight gradient has a simple
interpretation: it is the inner product between what token $i$ retrieves from
the update, $(K\Lambda)_i$, and its residual at the solution, $(V-KS)_i$.
From $\partial\mathcal{L}/\partial\bar S$, gradients flow further to the
retention gate via
$\partial\mathcal{L}/\partial\gamma_t
= \langle \Lambda, S_{t-1}\rangle_F$ and to the previous state via
$\gamma_t\Lambda$, continuing the recurrent chain across the TBPTT window;
the readout path contributes its own gradient to $S_{t-1}$ through
\cref{eq:read}. Gradients with respect to $w_i$ and the gate
pre-activations follow from $w_i=p_i\hat\beta_i/(1-\hat\beta_i)$ by the chain
rule. In practice we implement the forward and backward solves inside a
custom autograd function that stores $L$ from the forward pass, so that
\cref{eq:app-adjoint} costs one triangular solve pair rather than a
refactorization.

\section{Details of Real-World Experiments}
\label{appendix:setting of realworld}
\begin{figure}
    \centering
    \includegraphics[width=1.0\linewidth]{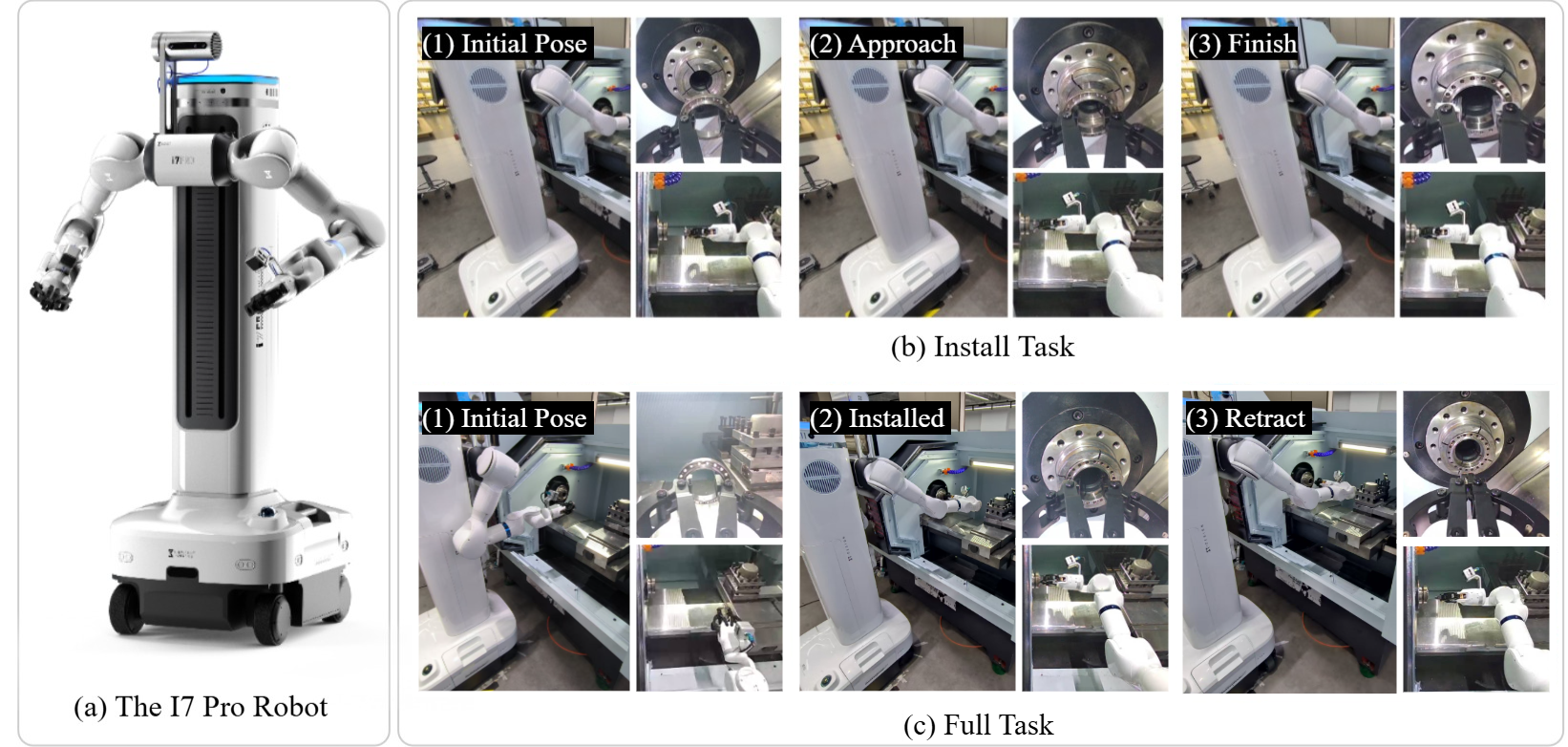}
    \caption{
Real-world experimental platform and manipulation tasks.
(a) The I7 Pro robot.
(b) The Install task, which requires precise alignment and insertion from a predefined pose near the target.
(c) The Full task, which extends Install with an initial approach, workpiece release, and arm retraction to complete the loading sequence.
For each task, three representative execution stages are illustrated using a global view, a close-up view of the manipulation, and a wrist-camera observation.
    }
    \label{fig:appendix:task_description}
\end{figure}

We conduct the real-world evaluation on the I7 Pro, a mobile manipulation platform equipped with dual robotic arms and onboard visual sensing, as shown in~\cref{fig:appendix:task_description}.
We consider two manipulation tasks in a CNC workspace: \emph{Install} and \emph{Full}.
In both tasks, the robot starts with the workpiece already grasped.
In the Install task, the arm begins at a predefined pose near the installation target and performs precise alignment and insertion.
The Full task extends this procedure into a complete loading sequence: the arm starts from a more distant initial pose, moves to the starting pose of the Install task, and completes the same alignment and insertion procedure.
After insertion, the robot opens the gripper to release the workpiece and retracts the arm.
Thus, Install focuses on fine manipulation near the target, while Full additionally requires the initial approach, workpiece release, and arm retraction.
For each task and method, we conduct 20 evaluation trials and report the success rate.
Representative execution stages and corresponding close-up views are shown in~\cref{fig:appendix:task_description}.

\section{Limitations}
Our evaluation considers episodic manipulation, with memory reset at episode boundaries. Memory management during continuous execution across successive tasks remains unexplored. Future work could investigate when to retain or reset memory as task instructions change.

\end{document}